\documentclass{article}

\usepackage{PRIMEarxiv}

\usepackage[utf8]{inputenc} % allow utf-8 input
\usepackage[T1]{fontenc}    % use 8-bit T1 fonts
\usepackage{hyperref}       % hyperlinks
\usepackage{url}            % simple URL typesetting
\usepackage{booktabs}       % professional-quality tables
\usepackage{amsfonts}       % blackboard math symbols
\usepackage{nicefrac}       % compact symbols for 1/2, etc.
\usepackage{microtype}      % microtypography
\usepackage{lipsum}
\usepackage{array, multirow}
\newcolumntype{L}{>{\centering\arraybackslash}m{3cm}}
\usepackage{fancyhdr}       % header
\usepackage{graphicx}       % graphics
\graphicspath{{media/}}     % organize your images and other figures under media/ folder

\usepackage{xcolor}
\definecolor{RoyalBlue}{RGB}{65, 105, 225}

\title{Using Large Language Models in Public Transit Systems: San Antonio as a Case Study
\thanks{\textit{\underline{Citation}}: 
\textbf{Authors. Title. Pages.... DOI:000000/11111.}} 
}

\author{
  Ramya Jonnala \\
  Texas A\&M University, San Antonio \\
  San Antonio, Texas\\
 \texttt{rjonn01@jaguar.tamu.edu} \\
   \And
  Gongbo Liang, Jeong Yang, Izzat Alsmadi \\
 Texas A\&M University, San Antonio \\
  San Antonio, Texas\\
\texttt{\{gliang, JYang, ialsmadi\}tamusa.edu} \\
}

\begin{document}
\maketitle

\begin{abstract}
The integration of large language models (LLMs) into public transit systems represents a significant advancement in urban transportation management and passenger experience. This study examines the impact of LLMs within San Antonio's public transit system, leveraging their capabilities in natural language processing, data analysis, and real-time communication. By utilizing GTFS and other public transportation information, the research highlights the transformative potential of LLMs in enhancing route planning, reducing wait times, and providing personalized travel assistance. Our case study is the city of San Antonio as part of a project aiming to demonstrate how LLMs can optimize resource allocation, improve passenger satisfaction, and support decision-making processes in transit management. We evaluated LLM responses to questions related to both information retrieval and also understanding. Ultimately, we believe that the adoption of LLMs in public transit systems can lead to more efficient, responsive, and user-friendly transportation networks, providing a model for other cities to follow.
\end{abstract}

% keywords can be removed

\section{Introduction}
The advent of artificial intelligence (AI) and machine learning (ML) has ushered in a new era of technological advancements that are transforming various sectors, such as cyber security~\cite{alsmadi2022adversarial,ahmad2022deep,liang2023enhancing}, healthcare~\cite{liang2019joint,xing2023self,liu2023simulated}, and public transportation~\cite{chen2021safety,alsrehin2023u2,liang2023unveiling}. Among these innovations, large language models (LLMs), such as OpenAI's GPT series~\cite{chatgpt2022,achiam2023gpt}, have demonstrated exceptional capabilities in natural language processing, understanding, and generation. These models can analyze vast amounts of data~\cite{kalla2023study,alyasiri2023exploring}, generate human-like text~\cite{orru2023human}, and facilitate complex decision-making processes~\cite{goossens2023integrating,li2023revolutionizing}, making them potentially invaluable tools for enhancing public transit systems. %This research paper explores the application of LLMs within the context of San Antonio's public transit system, aiming to illustrate their potential benefits and challenges.

Public transit systems are the backbone of urban mobility, providing essential services to millions of passengers daily~\cite{abdallah2023sustainable,xu2022urban}. Efficient and reliable public transportation is crucial for reducing traffic congestion, minimizing environmental impact, and promoting equitable access to mobility~\cite{guo2020systematic}. However, transit agencies often face challenges such as fluctuating passenger demand, route optimization, real-time communication with passengers, and efficient resource allocation~\cite{zhang2021agent,huang2020flexible}. Traditional methods of addressing these issues may fall short due to their limited scalability and adaptability.

San Antonio, one of the fastest-growing cities in the United States, presents a unique case study for examining the integration of LLMs in public transit. The city's rapid population growth has increased the demand for efficient public transportation solutions~\cite{altamirano2024to,houston2023san}.  The deployment of LLMs offers a promising avenue for addressing current challenges and future demands in public transportation as well as many other domains.

This study aims to investigate the potential of LLMs to improve various aspects of San Antonio's public transit system. Below are some of the potentials from employing LLMs in public transportation:
\begin{itemize}
    \item Optimize Route Planning and Scheduling: Evaluating how LLMs can analyze historical and real-time data to optimize routes and schedules, thereby reducing wait times and improving service reliability.
\item Enhance Passenger Communication: Exploring the use of LLMs for real-time interaction with passengers, providing personalized travel assistance, updates, and recommendations.
\item Improve Operational Efficiency: Assessing the impact of LLMs on resource allocation, including the deployment of buses and drivers, to enhance overall operational efficiency.
\end{itemize}

\section{Significance of the Study}
\label{sec:headings}

The integration of LLMs in public transit systems holds the potential to revolutionize urban mobility by making transportation more efficient, responsive, and user-friendly. This study not only contributes to the academic understanding of AI applications in transportation but also provides practical insights for transit authorities and policymakers. By focusing on San Antonio, a city representative of many growing urban areas, the findings can be generalized and applied to other cities facing similar challenges.

Furthermore, the research highlights the broader implications of AI in public services, emphasizing the importance of ethical considerations, data privacy, and the need for continuous evaluation and improvement. As cities worldwide grapple with the complexities of modern urbanization, the lessons learned from San Antonio's experience with LLMs can serve as a valuable guide for future innovations in public transit systems.

In conclusion, this study endeavors to bridge the gap between cutting-edge AI technologies and practical applications in public transportation, demonstrating how LLMs can be harnessed to create smarter, more adaptive, and passenger-centric transit networks. The following sections delve deeper into the theoretical framework, detailed methodology, findings, and implications of this transformative approach to public transit management.

\section{Related Work}
The integration of large language models (LLMs) like OpenAI's GPT-4 into public transit systems is a burgeoning field that aims to enhance the efficiency, accessibility, and user experience of public transportation. LLMs can process and analyze vast amounts of data, generate human-like text, and understand complex queries, making them suitable for a range of applications in public transit. This literature review explores the current state of research on the deployment of LLMs in public transit systems, focusing on areas such as passenger information services, operational efficiency, and accessibility improvements.

One of the primary applications of LLMs in public transit is in improving passenger information services. Studies have demonstrated that LLMs can enhance the quality and accuracy of real-time information provided to passengers. For instance, researchers explored the use of GPT in generating real-time updates and personalized travel advice for passengers, \cite{papangelis2020plato}, \cite{yenduri2024gpt}, \cite{voss2023bus}, \cite{khalil2024advanced}. Their findings indicated that LLMs could effectively handle complex passenger queries and provide accurate, context-aware responses, thereby improving the overall passenger experience.

Furthermore, researchers highlighted the potential of LLMs in multilingual support for transit systems, \cite{ullah2024role}, \cite{kaur2024text}, \cite{zheng2023trafficsafetygpt}. Given the diverse linguistic backgrounds of urban populations, LLMs like GPT-4 can be trained to provide information in multiple languages, ensuring that non-native speakers have equal access to transit information. This capability not only improves user satisfaction but also promotes inclusivity and accessibility.

The paper, \cite{devunurichatgpt} presents an evaluation of large language models (LLMs), specifically ChatGPT, in interpreting and retrieving information from General Transit Feed Specification (GTFS) data. The study demonstrates that ChatGPT can effectively understand and respond to various queries about public transit schedules and services, showcasing its potential in enhancing transit information systems. However, the paper also highlights areas for improvement, such as the model's occasional inaccuracies and the need for further fine-tuning to handle complex and domain-specific transit queries more reliably.

The paper, \cite{zheng2023chatgpt} explores the potential of using ChatGPT and similar large language models (LLMs) to revolutionize intelligent transportation systems. It argues that LLMs could significantly enhance various aspects of transportation, such as traffic management, passenger assistance, and operational efficiency, but also points out the challenges related to data privacy, model accuracy, and integration with existing systems.

\section{Goals and Approaches}
\label{sec:goal} 

% \gbl{Below are some quick writings. It might help us organize the manuscript and offer some ideas for the experiments. But we might need to update it later.}

% \gbl{Depending on how we narrate this manuscript, we might or might not need to mention the "dashboard" idea.}

Most LLMs today rely on learning-based methods. For example, the well-known ChatGPT~\cite{chatgpt2022} leverages the Transformer~\cite{vaswani2017attention} architecture and generative pre-training (GPTs)~\cite{radford2018improving,radford2019better,brown2020language,achiam2023gpt}. The output these models is inherently tied to the data they were trained on. Consequently, incorrect LLM responses can stem from multiple factors, such as limited information on a specific topic within the pre-training data or an LLM architecture (including its embedding method) incapable of correctly processing the user's input. Therefore, differentiating between pre-trained models and architectures is crucial when evaluating learning-based LLMs.

This project aims to assess LLMs' ability to understand GTFS (General Transit Feed Specification) and other public transportation information in two ways:
\begin{enumerate}
    \item \textbf{Performance of Common Pre-trained Models:} We will evaluate a pre-trained LLM model "as-is" by posing transportation-related questions and analyzing the accuracy of its responses. This assesses the model's ability to leverage its existing knowledge of GTFS data and public transportation information. Errors in this experiment might indicate either limited information within the pre-training dataset on the topic or an LLM architecture unsuited for handling the specific topic or questions. We denote this as the \textit{"understanding"} task in our experiments. 
    
    \item \textbf{Impact of LLM Architecture:} To delve deeper into the cause of errors, we propose a second experiment, assuming the LLM models have not encountered relevant information during pre-training. Before posing a specific transportation-related question, we will provide the necessary GTFS data and public transportation information to the LLMs and instruct them to answer based on the provided information. We will then re-ask the questions that resulted in failures during the first experiment. We denote this as the \textit{"information retrieval"} task in our experiments.
    
\end{enumerate}

The findings from these tasks will offer valuable insights into the cause of errors. For instance, if the LLMs can answer the questions correctly in the second experiment but not the first, it suggests insufficient pre-training data on the specific topic within the models. Conversely, the results might indicate that even with adequate data, the LLM models struggle with the questions, potentially due to architectural limitations.

% Chain of Thought prompting is a technique that guides Large Language Models(LLMs) through a step-by-step reasoning process to solve complex problems.It involves breaking down a task into sequential prompts, where each prompt build upon the previous one, gradually narrowing down the scope and providing additional context.This adaptive learning process allows the LLMs to maintain the context, build upon its previous reasoning, and deliver more accurate, coherent, and transparent responses.By structuring the thought process,Chain of Thought prompting enhances the logical reasoning and problem solving capabilities of LLMs, mitigating their limitations in these areas

\section{Experiments and Analysis}

\subsection{Experiment Setup}
This project specifically investigates the ability of LLMs to understand GTFS and public transportation information in the context of San Antonio's public transportation system. We leverage OpenAI's ChatGPT as the representative LLM due to its widespread public availability through both a web portal and a programmatic API. We designed a set of 275 questions specifically tailored to San Antonio's public transportation system. These questions are used to evaluate the LLM's performance in two key areas: 1) Understanding and 2) Information Retrieval (IR).

The Understanding task assesses how well the pre-trained ChatGPT model can comprehend and respond to questions about San Antonio's public transportation system (Goal \#1 in Section~\ref{sec:goal}). In contrast, the IR task examines the impact of LLM architecture on retrieving relevant information from a provided dataset (Goal \#2 in Section~\ref{sec:goal}). %By evaluating performance with access to relevant GTFS data, we aim to isolate the influence of the LLM architecture on task completion.

For our Understanding task, we employ 195 original multiple-choice questions (MCQs) with single correct answers that were meticulously crafted and span across the six question categories. The benchmarking dataset with all questionaires is made available to the public (see Appendix I). The breakdown of the number of questions in each category is presented in Table 1. We derived these questions and categories using the official GTFS Schedule documentation~\footnote{\href{https://gtfs.org/schedule/reference}{https://gtfs.org/schedule/reference}}.  
                                         
\begin{table}
    \centering
    \begin{tabular}{|c|c|} \toprule 
         Question Type& Number of Questions\\ \hline
         Term Definitions& 14\\ \hline
         Common Reasoning& 28\\ \hline
         File Structure& 17\\ \hline
         Attribute Mapping& 32\\ \hline
         Data Structure& 30\\ \hline
         Categorial Mapping& 74\\ \hline
         Total& 195\\ \hline
    \bottomrule
    \end{tabular}
    
    \caption{GTFS Understanding Benchmarking dataset questionnaire and their categories}
    \label{table:questions}
\end{table}

\begin{table}[!tb]
\centering
\begin{tabular}{ |c |c |c | p{10cm} |} \toprule 
 S.No& Category&  Type& Question \\ \hline
    
    1  & Categorial Mapping& Original & In the "trips.txt" file, what is the meaning of "wheelchair\_accessible" 0 or empty? a) No accessibility information for the trip b) Vehicle being used on this particular trip can accommodate at least one rider in a wheelchair c) No riders in wheelchairs can be accommodated on this trip d) Stop cannot be accessed by anyone A question \\ \hline
    2  & Attribute\ Mapping& Original& In which file does the shape\_dist\_traveled attribute\ appear in GTFS? a) stops.txt b) shapes.txt c) trips.txt d) stop\_times.txt A question \\ \hline
    3  & Common Reasoning& Original& Can a GTFS feed contain multiple agency information? a) Each agency should publish a seperate GTFS. b) No, GTFS feeds can only represent a single agency. c) Multiple agency information is specified in the "agency.txt" file. d) Agencies are not relevant in GTFS feeds. A question \\ \hline
    4& Data Structure& Original& How is the wheelchair\_accessible attribute represented in GTFS? a) Boolean (true or false) b) Float (number of accessible seats) c) Enum (e.g., 0,1,2) d) Text representation of wheelchair accessibility ...\\ \hline
 5&  File Structure& Original& What is the purpose of the "transfers.txt" file in GTFS? a) It contains information about fare rules and transfers. b) It provides details about the geographic shapes of routes. c) It specifies the frequency of trips. d) It provides real-time arrival and departure information. \\\hline
 6&  Term Definition& Original&What is a dataset in the context of GTFS? a) A single file containing all transit information b) A collection of tables representing different entities c) A specific date for transit service d) A record representing a transit agency \\ \hline
 7&  Attribute\ Mapping& Augmented& In which file can you find the route\_desc attribute in GTFS? a) stops.txt b) None of these c) trips.txt d) calendar.txt \\
 8&  Categorial Mapping& Augmented& What value is used in the "wheelchair\_boarding" field of the "stops.txt" file to indicate that the stop has no information regarding wheelchair accessibility? a) 0 b) 1 c) None of these d) 3 \\ \hline
 9&  Common Reasoning& Augmented& How does GTFS handle multiple trips on the same route at the same time? a) GTFS does not allow multiple trips on the same route at the same time. b) None of these c) Multiple trips are represented as separate routes in GTFS. d) GTFS relies on real-time updates to handle such cases. \\ \hline
 10& Data Structure& Augmented & What data type is used for the stop\_sequence attribute in GTFS? a) None of these b) Time c) Text d) Integer \\ \hline
 	\bottomrule
\end{tabular}
\caption{Ten questions that are used in this study. }
\label{table:questions}
\end{table}

In evaluating the LLM’s performance on MCQs, the model selects the answer (choice) with the highest probability for each question and output that without the need for any explanation. Although the LLM may always choose the correct answer when it is present, the LLM could opt for an alternate option when the correct choice is missing. To check the LLM’s robustness, we generate an augmented question set by creating variations of the original questions. Specifically, each original answer choice denoted as ‘a’, ‘b’, ‘c’, and ‘d’—is replaced one at a time with the phrase ‘None of these,’ resulting in additional 780 (195×4) variant questions and a total of 975 questions in the augmented dataset. The augmentation aims to evaluate how well the LLM can adapt to scenarios where the correct answer is removed.Refer to the above table for the examples of augmented questions

The ‘GTFS Retrieval’ benchmark employs a question-answer (QA) format, where no options are given and the LLM is supposed to give a single, correct answer. To prepare the questionnaire, we used the San Antonio VIA GTFS feed. The full feed included data on 98 bus routes. However, LLMs have limited context length: a metric for the number of tokens the LLM can process at once. The GPT-3.5-Turbo and GPT-4o have a maximum context length of 16,385 and 128k tokens respectively. The full GTFS feed is much larger than either LLM can accept, so we trim the dataset to just three bus routes (‘242’, ‘243’, and ‘246’) and 34 trips on these routes. These routes have 60 unique stops.

                   These questions in this benchmark range from basic lookup (single or multiple files) to performing data manipulations by the LLM. These include common data manipulation techniques like filtering, sorting, grouping, and joining. We divide the questions into two categories:
                   \begin{itemize}
                       \item Simple: These questions are based on simple lookups within the same file or two different files (using relational keys) within GTFS. Example: 
                What \texttt{route\_type} corresponds to \texttt{route\_id 243}?
                % – What is the trip\_id for the stop\_id 96746?
                % – What is the route\_long\_name for route\_id 242?
                % – How many unique stop\_id values are there for trip\_id 4430401?
                % – What is the departure\_time for the stop\_id 16689 and trip\_id 4430401?
 
                                       \item Complex : These questions need multiple files to extract information, require a deeper understanding, and could be open-ended. Example: 
                Tell the \texttt{route\_long\_name} in which there is a \texttt{stop\_name} as "GILLETTE \& PLEASANTON RD."? %\gbl{Ramya: could you add one or two sentences to explain what files are needed to support this question?} --Ans --- To support this question, the files needed are stops.txt, stop\_times.txt, trips.txt, routes.txt.To find the route long name, from the stop name provided can get the stopid from stops.txt  and using the retrieved stopid, can get the tripId from stop\_times.txt and based on that, can get the routeId from trips.txt and then from route Id, can get the route\_long\_name from routes.txt file.
                % – What is the route\_short\_name for the trips with shape\_id 211720?
                % - Tell the count of trips that visit the stop with \texttt{stop\_id 92797}?
                % – How many trips have both pickup\_type and wheelchair\_accessible as 1?
                % –What is the route\_long\_name for the route with the service\_id 293.0.1 and shape\_id 211720?
                   \end{itemize}
                   
% \gbl{Ramya: Could you also add a paragraph about how did you evaluate the model performance and what metric(s) was/were used?}

\subsection{Experimental Result}
\label{sec:result}
% \gbl{Suggestion on presenting the result: 1) show a few \textbf{interesting} examples, including both of the questions and answers. 2) report the quantitative evaluation results, e.g., the accuracy of each task.}

% \gbl{We might only report the results here and save the discussion \& analysis in a separate section.}

\begin{figure}
        \centering
        \includegraphics[width=1\linewidth]{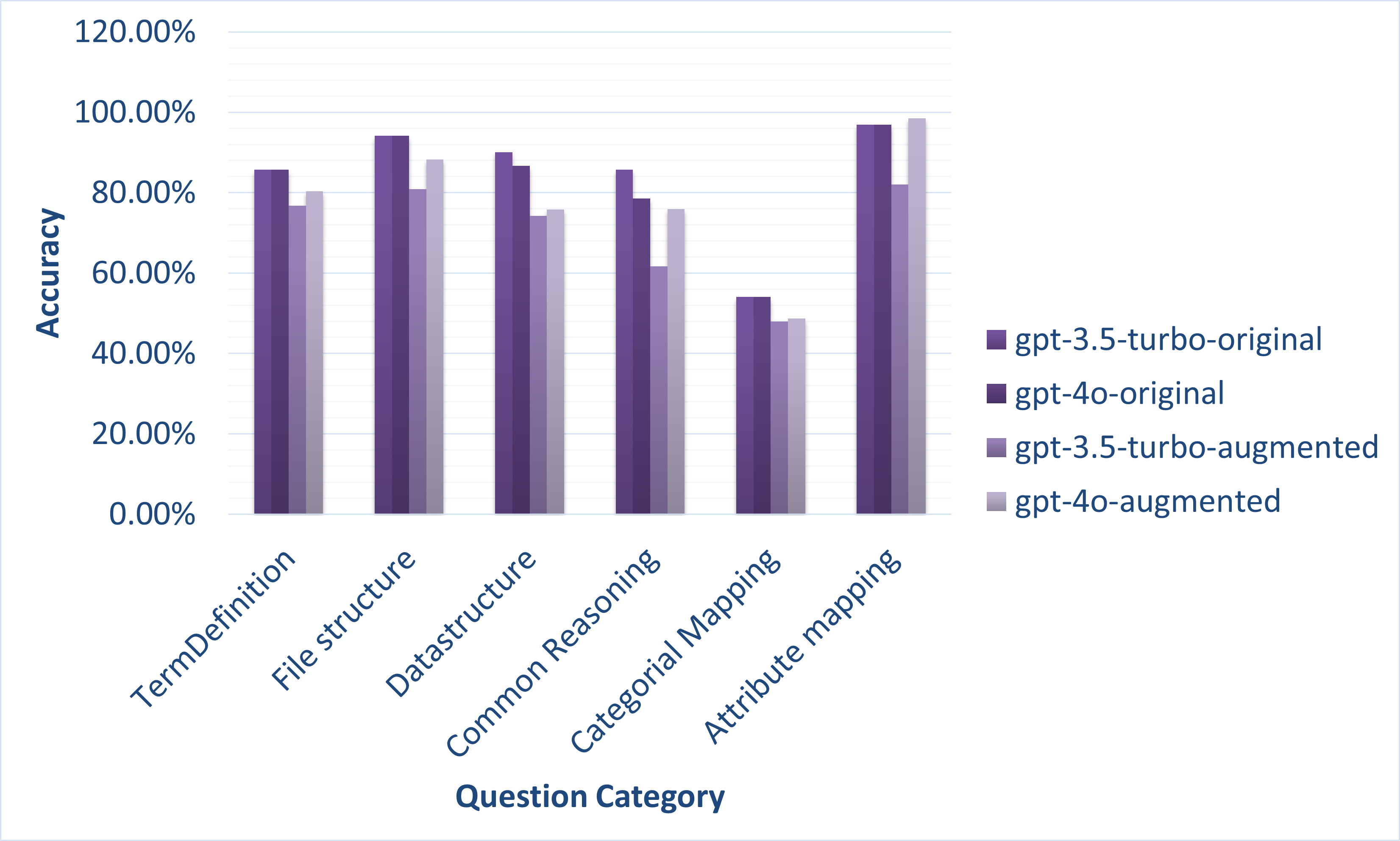}
        \caption{Summary of performance by question category for GPT-3.5-Turbo and GPT-4o on GTFS Understanding.}
        \label{fig:understanding}
\end{figure}

 In this study, we benchmarked both GPT-3.5-Turbo and GPT-4o on the original and augmented ‘GTFS Understanding’ dataset. Using the zeroshot learning (ZS) technique, the LLM attempts to answer the questions without been explicitly trained on. The accuracy of ZS on different categories of questions for both GPT-3.5-Turbo and GPT-4o is shown in Figure~\ref{fig:understanding}. The accuracy across both LLMs and all categories are higher on the original dataset than on the augmented dataset except for Attribute Mapping category. %~\gbl{Ramya: could you also include the performance numbers here? The figure only shows the tendency, but the exact number is harder to derived from there. --Ans --- Overall the accuracy is same in GPT-3.5-turbo and GPT-4o for the categories like Term Definitions with 85.71\%, 94.12\% in File Structure , with 54.05\% in Categorical Mapping and with 96.88\% in Attribute Mapping, but the accuracy in the categories like Data Structure and Common Reasoning is less in GPT-4o- with 86.67\% in DataStructure and 78.57\% in Common Reasoning than GPT 3.5-turbo - with 90.12\% in DataStructure and 85.71\% in Common Reasoning. } 
 This indicates that LLMs might not be robust to option substitution. But for the original dataset, the accuracy  of GPT-4o is equal or less than GPT-3.5-turbo. 
 %\gbl{Ramya: this is very interesting! People would expect GPT-4o has equal or better performance. So, could you show the example that GPT-4o got wrong here? Ideally, we need to offer some opinions about why this happened. -- Ans --For the example shown below is where the GPT-4o got wrong            -How is the email attribute represented in GTFS? a) Email address b) Enum c) Phone Number d) URL The answer got from GPT-4o is d whereas from GPT-3.5-turbo is a. So as mentioned above, the two categories i.e DataStructure and CommonReasoning related questions accuracy is lower in GPT-4o than GPT-3.5-turbo}
 
 The discussions in the remainder of the paper are focused to the augmented dataset alone. Overall, GPT-4o performs better than GPT-3.5-Turbo, with above 88\% accuracy in “File Structure”, above 98\% accuracy in "Attribute Mapping, above 80\% in “Term Definitions”, and above 75\% accuracy in  “Data Structure” and “Common Reasoning”. However, GPT-4o and GPT-3.5-turbo achieve a below 50\% accuracy for “Categorical Mapping”. % ~\gbl{Rayma: could you try to offer some reasons about why this happened? -- Ans --Since LLMs embed words in latent space, the embedding for categorical variables could be similar, making it difficult for the LLMs to distinguish and easier to hallucinate. Take, for example, the categories ‘Rail’ and ‘Light Rail’ in the \texttt{route\_type} attribute. The words ‘Rail’ and ‘Light Rail’ have a cosine similarity of 0.88. However, in the context of GTFS, both have different meanings. The ‘Rail’ has \texttt{route\_type} of 2, and ‘Light Rail’ has \texttt{route\_type} of 0.} 
 The GPT-3.5-Turbo has around 70\% accuracy for all categories, except “Categorical Mapping”, which has the worst accuracy for both LLMs. 

\begin{figure}
    \centering
    \includegraphics[width=0.75\linewidth]{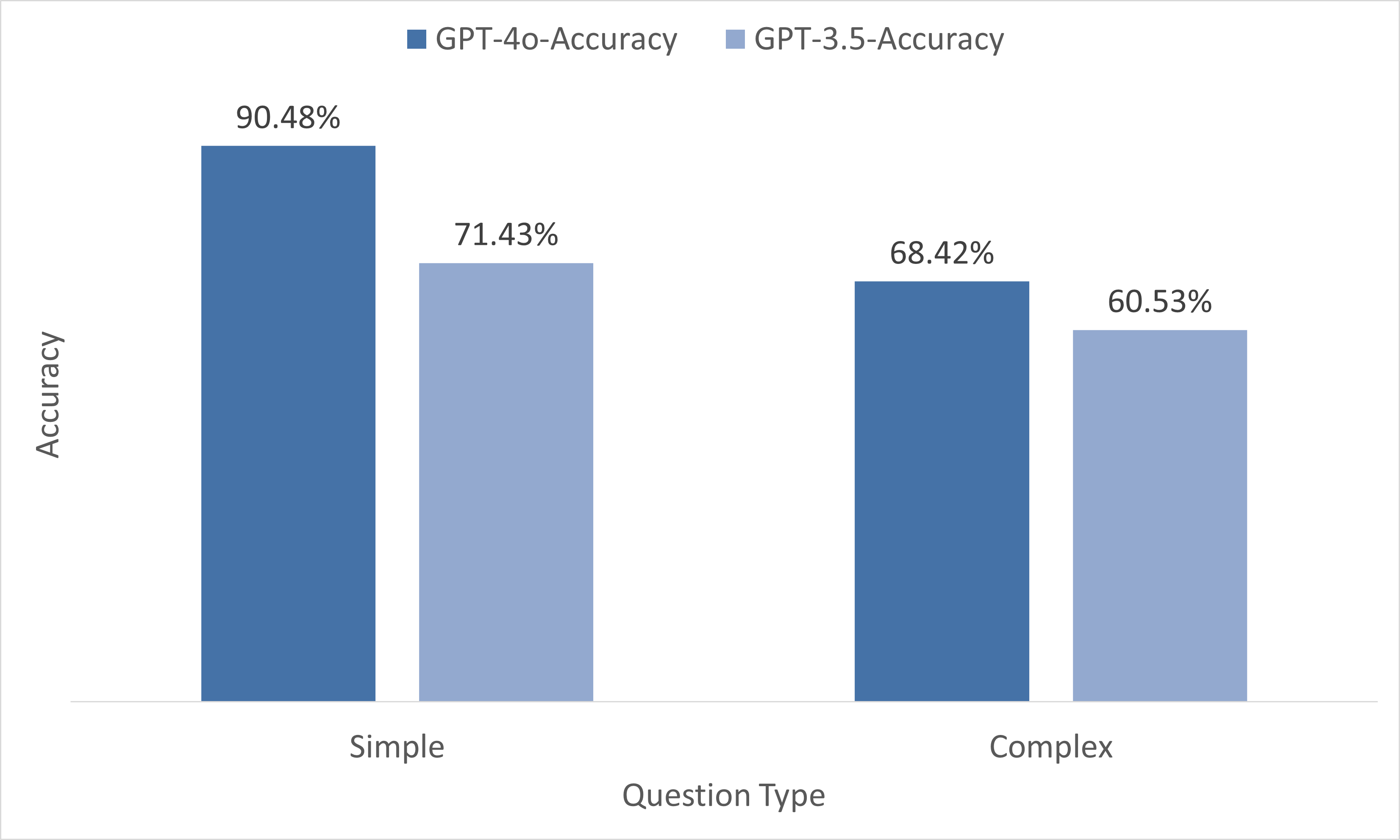}   
    \caption{Summary of performance by question type for GPT 3.5-turbo and GPT-4o on GTFS Retrieval Benchmark}
    \label{fig:retrieval}
\end{figure}

Similar to testing the understanding of GTFS, we pose questions to the LLM to see its capabilities in information retrieval. A total of 80 questions were posed with 42 simple and 38 complex questions. Using the ZS technique, posed these questions. Before posing these questions, extracted the content from all the files of the filtered data. %~\gbl{Rayma: how did you conduct the retrieval task is a little unclear here. Basically, you extract the contents from the file, upload the contents to GPTs, then asked the question. Right? Could you refine this part a little bit? Also, I don't think if we have mentioned filtered data yet. So you might want to explain that too. -- Ans --I have extracted the content from each file and then appended all the content from all the files.Then I passed the final content and questions to the messages parameter chatCompletions API} 
The extracted content and questions were posed to the gpt API. The results in the Figure~\ref{fig:retrieval} shows that the accuracy of gpt-4o is significantly better than gpt-3.5-turbo

For simple type question, the accuracy of gpt-4o is \~15\% higher than gpt-3.5-turbo and for the complex type question, the accuracy of gpt-4o is \~8\% higher than gpt-3.5-turbo. The overall perfomance of gpt-4o in the IR task is very much better than gpt-3.5-turbo model

% \subsection{Analysis and Discussion}

\section{Conclusion and Future Work}
This work evaluates the ability of Large Language Models (LLMs) to understand public transportation information through two tasks: \textit{"understanding"} and \textit{"information retrieval."} The LLMs achieved accuracy ranging from $47.97\%$ to $98.44\%$ on the understanding task and $60.53\%$ to $90.48\%$ on information retrieval. The high performance on some understanding tasks suggests that pre-trained LLM models have acquired a significant amount of transportation-related information from their training datasets. However, the large gap between the best and worst performing tasks also indicates that the models might have been trained on an imbalanced dataset, with significantly less information on certain areas. While relevant information is given, modern LLM models can handle task about to unknown data, suggested by the high performance on the information retrieval task. However, their ability to do so seems to be significantly reduced when the task complexity increases.  

This work demonstrated the use of large language models in public transit systems holds great promise for transforming how these systems operate and serve their users. From improving passenger information services and operational efficiency to enhancing accessibility, LLMs offer a wide range of applications that can significantly benefit public transit. However, the large performance gaps between the best and worst performing tasks needed to be address before using it in the real-world. In addition, addressing ethical concerns and ensuring the responsible use of these technologies will be essential as this field continues to evolve. With continued research and development, LLMs have the potential to play a pivotal role in the future of public transportation.
\section{Acknowledgment}
This work is supported by the National Science Foundation under Grant No. 2131193. Any opinions, findings, conclusions, or recommendations expressed in this material are those of the author(s) and do not necessarily reflect the views of the National Science Foundation

%Bibliography
\bibliographystyle{unsrt}  
\bibliography{references}

\end{document}